\pdfoutput=1

\documentclass[11pt]{article}

\usepackage{EMNLP2022}

\usepackage{times}
\usepackage{latexsym}

\usepackage[T1]{fontenc}

\usepackage[utf8]{inputenc}

\usepackage{microtype}
\usepackage{graphicx}

\usepackage{inconsolata}
\usepackage{MnSymbol}
\usepackage{multirow}
\usepackage{colortbl}
\usepackage{color}

\title{Grafting Pre-trained Models for Multimodal Headline Generation}

\author{
    Lingfeng Qiao$^\dag$, Chen Wu$^\dag$, Ye Liu$^\dag$, Haoyuan Peng$^\dag$, Di Yin$^\dag$, Bo Ren$^\S$ \\ $^\dag$Tencent Youtu Lab, Shanghai, China \\ $^\S$Tencent Youtu Lab, Hefei, China \\ \texttt{\{leafqiao,rafelliu,haoyuanpeng,endymecyyin,timren\}@tencent.com}, \texttt{overwindows@icloud.com}}
    
\begin{document}
\maketitle
\begin{abstract}
Multimodal headline utilizes both video frames and transcripts to generate the natural language title of the videos.
Due to a lack of large-scale, manually annotated data, the task of annotating grounded headlines for video is labor intensive and impractical.
Previous researches on pre-trained language models and video-language models have achieved significant progress in related downstream tasks. 
However, none of them can be directly applied to multimodal headline architecture where we need both multimodal encoder and sentence decoder.
A major challenge in simply gluing language model and video-language model is the modality balance, which is aimed at combining visual-language complementary abilities.
In this paper, we propose a novel approach to graft the video encoder from the pre-trained video-language model on the generative pre-trained language model.
We also present a consensus fusion mechanism for the integration of different components, via inter/intra modality relation.
Empirically, experiments show that the grafted model achieves strong results on a brand-new dataset collected from real-world applications.

\end{abstract}

\section{Introduction}

\begin{figure}[t]
  \centering
  \includegraphics[width=0.99\linewidth]{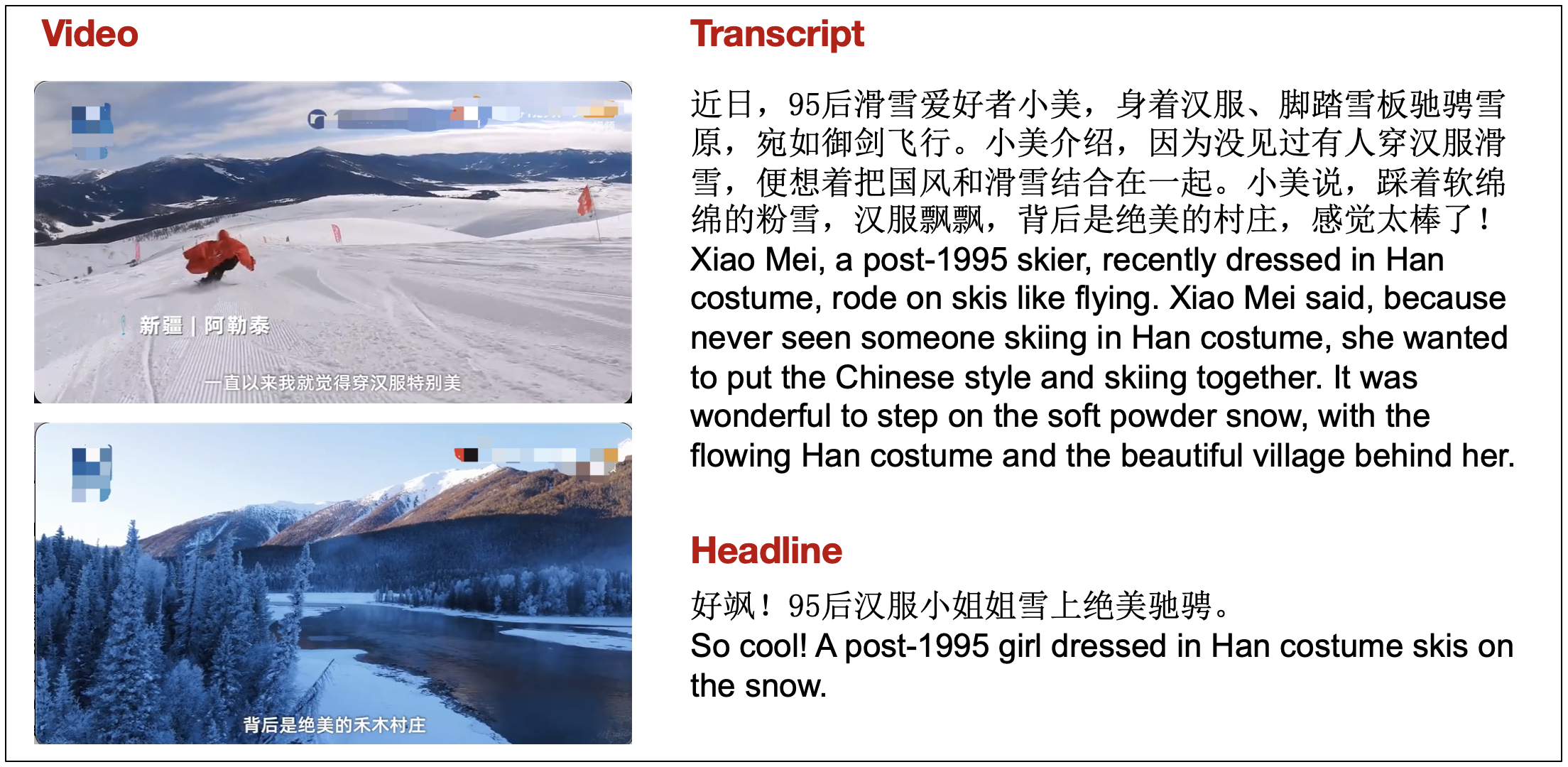}
  \caption{An example of multimodal generation.}
  \label{FIG: video summary definition}
  \vspace{-2ex}
\end{figure}

In the age of information explosion, generating headlines of videos has been steadily gaining prominence on the short video platform.
As the headlines can summarize the videos for people quickly acquiring their essential information. Good headlines are also beneficial for various scenarios, such as video retrieval, recommendation, and understanding \cite{zhu2022leveraging, liu2022msl}.
Specially, video headline generation can be regarded as a textual generation task with multimodal inputs \cite{li2021pretrained}, which is called multimodal generation as shown in Figure \ref{FIG: video summary definition}. 
Given a video with related transcript, algorithm aims to generate a short, concise and readable textual attraction title. 


However, to build an effective model, collecting large-scale training data is the main challenge. 
Especially, in the multimodal headline generation task, the form of triplet data $($video, source transcript, target summary$)$ further increases the difficulty of collecting data, which limits the application of video headline generation.

To alleviate the issue in multimodal headline generation, the natural idea is to leverage pre-trained model (PTM).
With large-scale corpus, such as GPT \cite{radford2018improving}, BART \cite{lewis2019bart}, PALM \cite{bi2020palm}, etc., have shown great ability to generate readable and informative text. 
Since the multimodal headline generation combine both video and textual information, we propose that the model can be grafted by PTMs of language generation and video-text matching. The former provides the ability of headline generation, and the latter bridges the semantic gap between the multiple modalities \cite{radford2021learning}. In addition, these two tasks have no concern of scarce data. 
For language generation, the existing pre-training model can be directly adopted. For video-text matching, the model can be trained with sufficient data from the Internet without manual annotations. 
By this means, multimodal headline generation model can be constructed by fine-tuning the grafted model with limited data collection.


In order to take advantage of the existing PTMs and improve reusability, we propose a \textbf{gra}fting mechanism for obtaining the \textbf{m}ultimodal summarization pre-training \textbf{mo}del (GraMMo). First, language generation and video-text matching tasks are introduced to pre-train the encoders and decoder, respectively. 
Then we construct a unified architecture with a video encoder, a text encoder and a text decoder which grafted from different PTMs to initialize the multimodal headline generation model.
In addition, a joint-modality layer acted as a modality-balance gate is designed to fuse the video and text features.
Unlike previous works which focus on retaining modality-shared feature \cite{libovicky2018multimodal, yu2021vision}, this layer uses a two-way attention strategy to capture the commonality and specialty of the modalities. 
In detail, the video modality can highlight the most relevant and important text tokens by video-text cross attention. The resulted feature is called video-enhanced text feature, which reflect the commonality. On the other hand, since video-enhanced text feature neglects the video specialty which is less related with text modality, we recombine the video embeddings according to video-text attention and exploit the video-specific feature for complementing the fusion representation. Furthermore, dynamic frame sampling (DFS) and masked word prediction are designed in the encoder parts to reinforce the multimodal representation.
We summarize the main contributions as follows:

\begin{itemize}
\item We propose a grafting video-text pre-training framework for multimodal headline generation. By grafting PTMs of language generation and video-text matching, GraMMo can be efficiently trained without big data collection. It is beneficial for fast deployment of real-world applications.


\item A joint-modality layer with multimodal fusion module is designed to pay balanced attentions to each modality. It uses a two-way attention strategy to capture the commonality and specialty of multiple modalities, which will reinforce the fusion representation for better headline generation.

\item Extensive experiments on a proposed Chinese multimodal headline generation dataset, WB-News, demonstrate that the grafted model can effectively accelerate the downstream fine-tuning procedure and improve generation results. The proposed method has been deployed in an industrial media platform for Chinese video headline generation.




\end{itemize}

\begin{figure*}[t]
  \centering
  \includegraphics[width=0.99\linewidth]{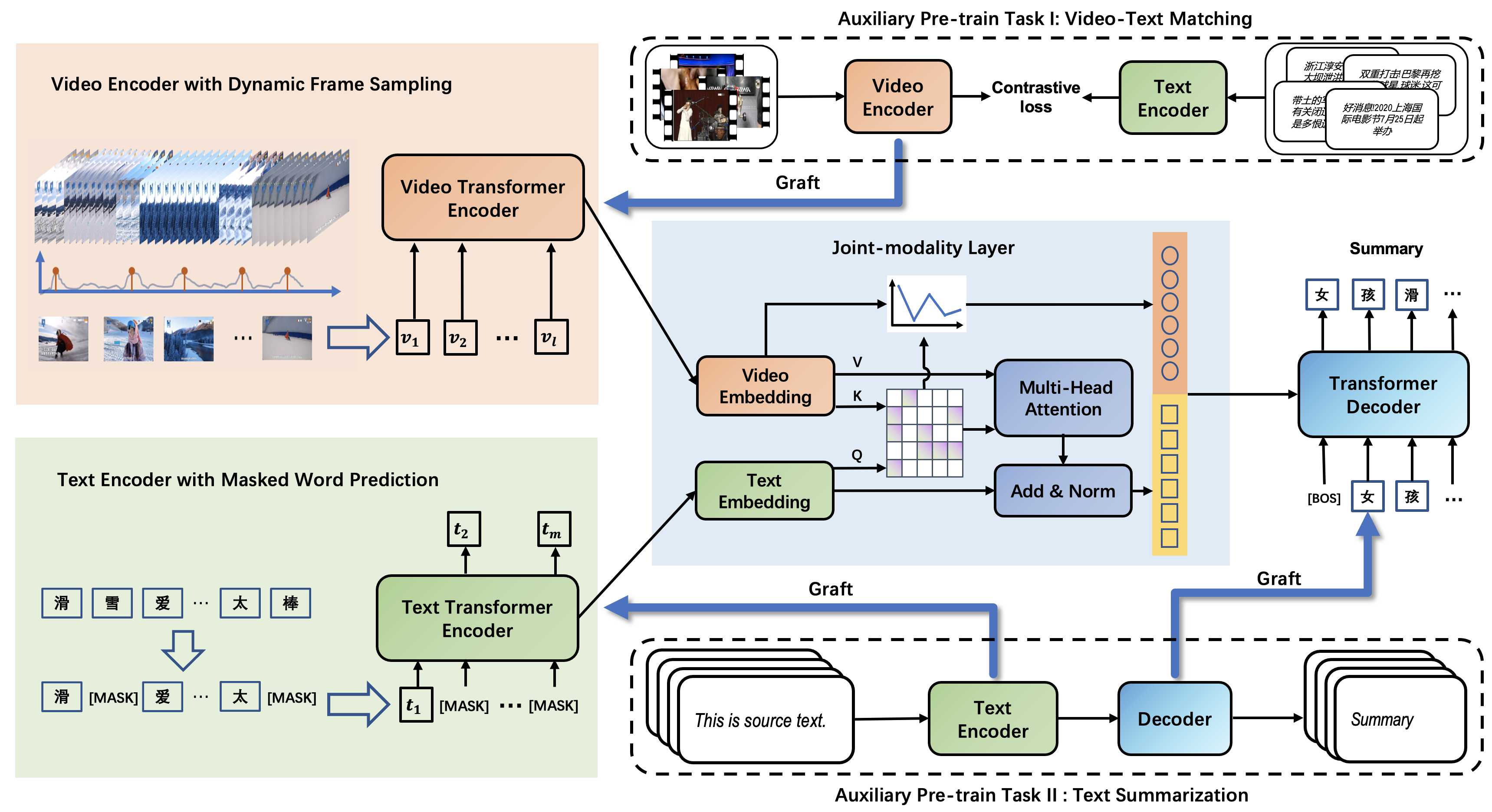}
  \caption{GraMMo framework for multimodal headline generation. }
  \label{FIG: grafting framework}
\end{figure*}

\section{Related Work}

\subsection{Multimodal Generation}
Multimodal generation task aims to generate short, concise and readable textual title that can capture the most core information of the input media. The task is closely relevant to text summarization \cite{zhang2020pegasus, jiang2022leveraging} while it is much tougher because redundancy and complementary between multiple modalities should be studied \cite{jangra2021survey}. Many literatures have been based on pre-extracted unimodal sequential representation and cross attention mechanism to obtain the fused feature \cite{li2020vmsmo, khullar2020mast, li2020multimodal, fu2020multi}. Besides, some researchers took efforts to sufficiently fusing the multimedia inputs by hierarchical fusion \cite{liu2020multistage, yu2021vision, zhang2021hierarchical}. The objective of modality consistency is another tool to guide the learning of multimodal fusion \cite{zhu2020multimodal, zhang2021unims}.


However, few existing methods studied PTM for multimodal generation \cite{seo2022end}. In this paper, we propose a grafting video-text generation model and a novel joint-modality layer which is designed to capture the commonality and specialty of multiple modalities.




\subsection{Video-Text Pre-training}


Video-Text pre-training models adopt the "pre-training and then fine-tuning" paradigm, which makes the downstream tasks able to utilize the abundant knowledge included in pre-training data. 

One class of work is task-specific pre-training, and contrastive learning is used for zero-shot transfer and video-text retrieval tasks, such as CLIP \cite{radford2021learning} and other related researches \cite{miech2019howto100m, patrick2020support, huang2021multilingual, xu2021videoclip}. 
Furthermore, CBT \cite{sun2019learning}, HERO \cite{li2020hero}, VideoAsMT \cite{korbar2020video} and UniVL \cite{luo2020univl} adopt multi-task learning (MTL) for pre-training on retrieval tasks. 
The other class of work concentrates on how to interact the multimodal inputs, including VideoBERT \cite{sun2019videobert}, Unicoder-VL \cite{li2020unicoder}, VL-BERT \cite{su2019vl}, UNITER \cite{chen2020uniter}, VLP \cite{zhou2018end}, ActBERT \cite{zhu2020actbert}, VLM \cite{xu2021vlm} and BEiT \cite{wang2022image}.



Currently, few video-text pre-training models focus on multimodal headline generation due to the scarce data. GraMMo gives an effective grafting architecture for this task with ready-made PTMs of language generation and video-text matching, which can save a lot of computational costs.

\section{Approach}

\subsection{Grafting Architecture}
Given an input video $X_{v}=\{v_{1}, v_{2}, \cdots, v_{l}\}$ and related source transcript $X_{t}=\{t_{1}, t_{2}, \cdots, t_{m}\}$, the output is the target textual title $Y=\{y_{1}, y_{2}, \cdots, y_{n}\}$, where $l, m, n$ are the numbers of the corresponding tokens.
The goal is to generate a predicted title $Y'=\{y'_{1}, y'_{2}, \cdots, y'_{n}\}$ based on $X_{v}$ and $X_{t}$, which can successfully grasp the main points of the video and transcript.

The proposed architecture is to provide a PTM for multimodal headline generation without large-scale triplet samples $\{X_{v},X_{t},Y\}$. The concept of grafting architecture GraMMo is illustrated in Figure \ref{FIG: grafting framework}. As a unified structure for multimodal generation, GraMMo consists of a video encoder $E_{v}(\cdot)$, a text encoder $E_{t}(\cdot)$, a joint-modality layer $F(\cdot)$ and a text decoder $D(\cdot)$. The encoders are designed for each modality individually and the architecture can be easily extended to various multimodal tasks with different kinds of inputs. Then the embeddings of modalities are fused by joint-modality layer to obtain the multimodal features. The joint-modality layer can provide video-enhanced text feature and video-specific feature, which involve the commonality and speciality of video and text modalities. Finally, text decoder is used to generate headline based on the fused multimodal feature.


To pre-train the video encoder, text encoder and text decoder, we draw support from two auxiliary tasks, i.e. language generation and video-text matching. As Figure \ref{FIG: grafting framework} shows, the video encoder of video-text matching, the text encoder, and the text decoder of language generation are grafted to obtain the multimodal headline generation model.

\subsection{Pre-train Generative Language Model}

The language generation model with encoder-decoder structure can be adopted to build text encoder and decoder. In the work, PALM \cite{bi2020palm} which is Transformer-based \cite{vaswani2017attention} architecture is used as the NLG model. The text encoder and text decoder are pre-trained as classic abstractive text summarization task with large-scale unlabeled corpus.



In the pre-training stage, text encoder $E_{t}(\cdot)$ encodes source transcripts to obtain text embeddings $e_{t}=E_{t}(X_{t})$, and then the decoder $D(\cdot)$ learns to generate hypothesis summaries. The generation loss and masked word prediction loss are used to guide the learning of text encoder and text decoder.






\subsection{Pre-train Video-Text Understanding}

We also use Transformer architecture for video encoder. The video features, extracted by I3D network \cite{carreira2017quo}, are first projected to video tokens before being fed into the video Transformer. For a video $X_{v}$, it has $s$ frames, which is denoted as $V_{f}=\{f_{1},f_{2},\cdots,f_{s}\}$. Due to the concern of model efficiency, the frames should be sampled and converted to the video tokens with $l$ length. Most conventional methods used pre-extracted features based on uniform sampling from the raw video, e.g. extract one frame every one second duration of the video. 
However, this sampling method may limit the expression of video. In the framework, to enhance video embeddings, dynamic frame sampling (DFS) is designed. It is a projection layer $DFS(\cdot):\{1,2,\cdots,l\}\to \{1,2,\cdots,s\}$, which represents the choice from variable frames.


As a result, the video tokens can be obtained by $v_{i}=f_{DFS(i)}$, where $i=1,2,\cdots,l$. Then the I3D features of these tokens are extracted and video embeddings $e_{v}=E_{v}(X_{v})$ are acquired by a stacked Transformer encoder.

To pre-train the video encoder $E_{v}(\cdot)$, video-text matching task is adopted to bridge the semantic gap between video and text modalities. Specifically, we collect large-scale video-text pairs from public video platform on the Internet without manual annotations. Videos and their corresponding descriptions are the natural data for video-text matching. The proposed video encoder and another Transformer-based text encoder are used to encode videos and descriptions respectively. Contrastive loss InfoNCE \cite{oord2018representation} is employed to calculate the correspondence between embeddings and guide the pre-training of encoders.

\subsection{Joint-Modality Layer}


Joint-modality layer will be used in fine-tuning stage after model grafting. 
Given text embeddings $e_{t}\in R^{m\times d}$ and video embeddings $e_{v}\in R^{l\times d}$, where $d$ is the dimension of the embeddings, joint-modality layer is proposed to fuse them for headline generation. First, video embeddings should be used to highlight the significant elements in text embeddings and make algorithm pay attention to them from redundant source transcripts. Second, video embeddings can supplement key information that is not included in text embeddings to improve the informativeness of headline. To realize these motivations, as Figure \ref{FIG: grafting framework} shown, the joint-modality layer uses a two-way attention strategy to capture the commonality and speciality of multiple modalities. Denote the query $Q\in R^{m\times d}$ is projected from text embeddings $e_{t}$, and the key $K\in R^{l\times d}$ and value $V\in R^{l\times d}$ are projected from video embeddings $e_{v}$. With dot-product between $Q$ and $K$, the video-text attention matrix $M_{vt}\in R^{m\times l}$ is obtained, which represents the relations between text and video tokens. On the one hand, the text features are enhanced by video information based on $M_{vt}$. Multi-head attention is applied and $V$ is added to the related text tokens to obtain the video-enhanced text feature $g=e_{t}+M_{vt}V$.



On the other hand, video embeddings $e_{v}$ can be divided into two aspects, i.e. text-relevant feature and video-specific feature. Text-relevant feature is the part of $e_{v}$ with large video-text attention score and video-specific feature is the opposite part. The text-relevant feature has already been considered in $M_{vt}V$. On the contrary, the video-specific feature is neglected and should be supplemented. We calculate video-text relevant distribution $p\in R^{1\times l}$ by summing $M_{vt}$ along the query dimension. The lower value in $p$ means that such a video embedding is less relevant to text, which should be chosen for video-specific feature. As a result, the video-specific feature $h$ is obtained by $h=e_{v}\odot Norm(1-p)$, where $Norm(\cdot)$ means the normalized operator and $\odot$ is the element-wise multiplication.



Finally, the fusion embeddings $F(e_{t},e_{v})$ is obtained by concatenating video-enhanced text feature $g$ and video-specific feature $h$.




\subsection{Fine-tune Multimodal Generation Model}


As mentioned above, text encoder $E_{t}(\cdot)$ and text decoder $D(\cdot)$ are pre-trained by language generation task. Video encoder $E_{v}(\cdot)$ is pre-trained by video-text matching task. By grafting these modules with joint-modality layer, a multi-modality generation model is established, which can be used for multimodal headline generation task.

For realistic applications, the specific multimodal headline generation triplet data should be collected to fine-tune the model. With GraMMo, we only need to prepare a small amount of data, since most of parameters in model are initialized by grafting, the model can converge rapidly.

\begin{table*}[ht]
\small

\centering 
\begin{tabular}{p{6cm} | c | c | c | c | c | c | c | c} 
\hline 
\hline
\textbf{Methods} & \textbf{R-1} & \textbf{R-2} & \textbf{R-L} & \textbf{B-1} & \textbf{B-2} & \textbf{B-3} & \textbf{B-4} & \textbf{M} \\ 
\hline 
\hline
\rowcolor{green!20!blue!10} \emph{Text ---> Text} & & & & & & & &\\
BART \cite{lewis2019bart} & 33.14 & 19.51 & 29.41 & 32.54 & 25.34 & 19.60 & 15.39 & 29.48\\ 
PALM \cite{bi2020palm} & 36.73 & 23.10 & 33.86 & 34.01 & 27.47 & 21.76 & 17.37 & 32.10\\
\hline
\hline
\rowcolor{green!20!blue!10} \emph{Video ---> Text} & & & & & & & &\\
VLM \cite{xu2021vlm} & 5.10 & 0.61 & 4.44 & 4.31 & 1.54 & 0.64 & 0.29 & 2.89\\
GraMMo-Video & 7.85 & 1.73 & 6.90 & 7.21 & 3.31 & 1.76 & 1.01 & 5.06\\ 
\hline
\hline
\rowcolor{green!20!blue!10} \emph{Video+Text ---> Text} & & & & & & & &\\
VG-GPLM \cite{yu2021vision} & 35.35 & 21.46 & 32.10 & 33.81 & 26.70 & 20.78 & 16.38 & 31.01 \\
MV-GPT \cite{seo2022end}  & 37.74 & 24.04 & 34.42 & 36.13 & 29.27 & 23.34 & 18.81 & 33.73 \\
MMPT \cite{xu2021videoclip} & 38.21 & 24.25 & 35.05 & 31.65 & 25.77 & 20.58 & 16.65 & 31.76 \\
\textbf{GraMMo} & \textbf{38.87} & \textbf{24.85} & \textbf{35.38} & \textbf{37.80} & \textbf{30.65} & \textbf{24.43} & \textbf{19.65} & \textbf{35.30} \\ 
\hline 
\hline
\end{tabular}
\caption{Headline generation results on WB-News dataset.} 
\label{table:summarization result 1} 
\end{table*}

\section{Experiments}
\subsection{Datasets and Implementation}

We leverage billions of Chinese corpus and millions of videos for pre-training language model and video-text matching models, respectively. 
For multimodal headline generation fine-tuning and testing, a new dataset WB-News is built.
The details of datasets and the methodology used to obtain the corpus can be found in supplementary materials \ref{sec:experimental}.
For evaluation on WB-News dataset, three metrics are employed: ROUGE (R-1,R-2,R-L) \cite{rouge2004package}, BLEU (B-1,B-2,B-3,B-4) \cite{papineni2002bleu}, METEOR (M) \cite{banerjee2005meteor}.


\subsection{Headline Generation Results}
On WB-News dataset, according to different types of input, we compare generation methods in three kinds of experimental setups, i.e. \textit{Text ---> Text}, \textit{Video ---> Text} and \textit{Video+Text ---> Text}. \textbf{BART} and \textbf{PALM} are used as baselines for classic text-only generation. 
Without text modality, \textbf{GraMMo-Video} is designed by pruning text encoder in GraMMo, which is compared with the video caption method \textbf{VLM}.
For multimodal generation task, latest methods \textbf{VG-GPLM}, \textbf{MV-GPT} and \textbf{MMPT} are compared with the proposed \textbf{GraMMo}. 
As Table \ref{table:summarization result 1} shows, GraMMo achieves the best results among related multimodal generation methods and the SOTA text summarization method, PALM, with large margin. 
It illustrates that the video modality can help text to improve the headline results and GraMMo can better leverage the multimodal information against the related methods.

In the \textit{Video ---> Text} scenario, because the factual information, such as the name, cannot be extracted using the video modality, the headline metrics are quite low. Nevertheless, \textbf{GraMMo-Video} achieves a better performance and can generate reasonable summaries if neglecting the factual consistency. It also shows that our method has the ability to extract the language semantics from video. 

Furthermore, GraMMo has been deployed on an industrial platform for video headline generation, which is shown in the discussion section.

\subsection{Grafting for Headline Generation}

\begin{table}[t]
\small
\centering 
\begin{tabular}{p{1.3cm} | p{1.3cm} | c | c | c} 
\hline 
\hline
\multicolumn{2}{c|}{\textbf{Pre-trained Models}}    &  \multicolumn{3}{c}{}  \\ \hline
\multicolumn{1}{c|}{\textbf{Text}} & \multicolumn{1}{c|}{\textbf{Video}} & \textbf{R-L} & \textbf{B-4} & \textbf{M} \\ \hline
\hline 
 &  & 26.98 & 12.18 & 24.65 \\ \hline
  & \multicolumn{1}{c|}{$\checkmark$} & 27.63 & 12.30 & 24.69\\ \hline
 \multicolumn{1}{c|}{$\checkmark$} &  & 34.90 & 18.63 & 34.06\\ \hline
 \multicolumn{1}{c|}{$\checkmark$} &  \multicolumn{1}{c|}{$\checkmark$} & \textbf{35.38} & \textbf{19.65} & \textbf{35.30}\\
\hline 
\hline
\end{tabular}
\caption{The effectiveness of using grafted model.} 
\label{table:summarization result 2} 
\end{table}

\begin{figure}[t]
  \centering
  \includegraphics[width=0.7\linewidth]{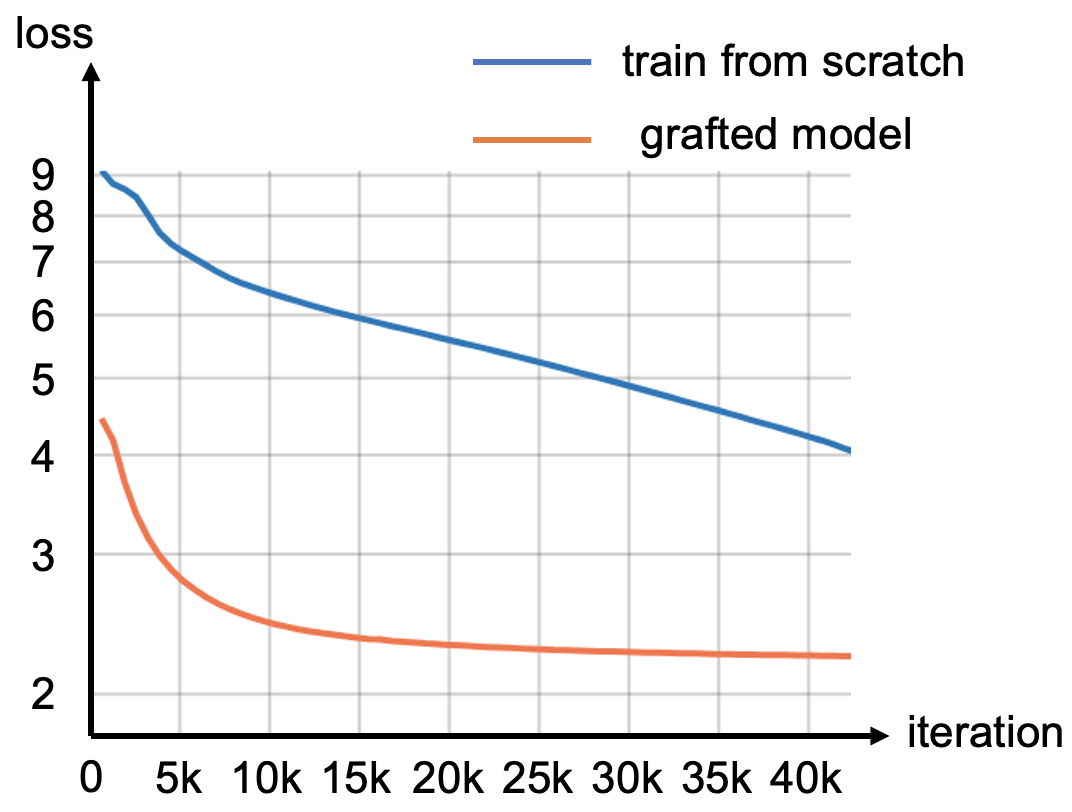}
  \caption{Different initialization methods.}
  \label{FIG: learning curves}
\end{figure}

In Table \ref{table:summarization result 2}, the generation results are compared by eliminating grafted components. 
We found that the performance can be significantly improved by grafting. 
Figure \ref{FIG: learning curves} also shows that the learning curve of fine-tuning with grafted model decreases and convergences more rapidly against the case of train-from-scratch. 

\begin{figure*}[ht]
  \centering
  \includegraphics[width=0.99\linewidth]{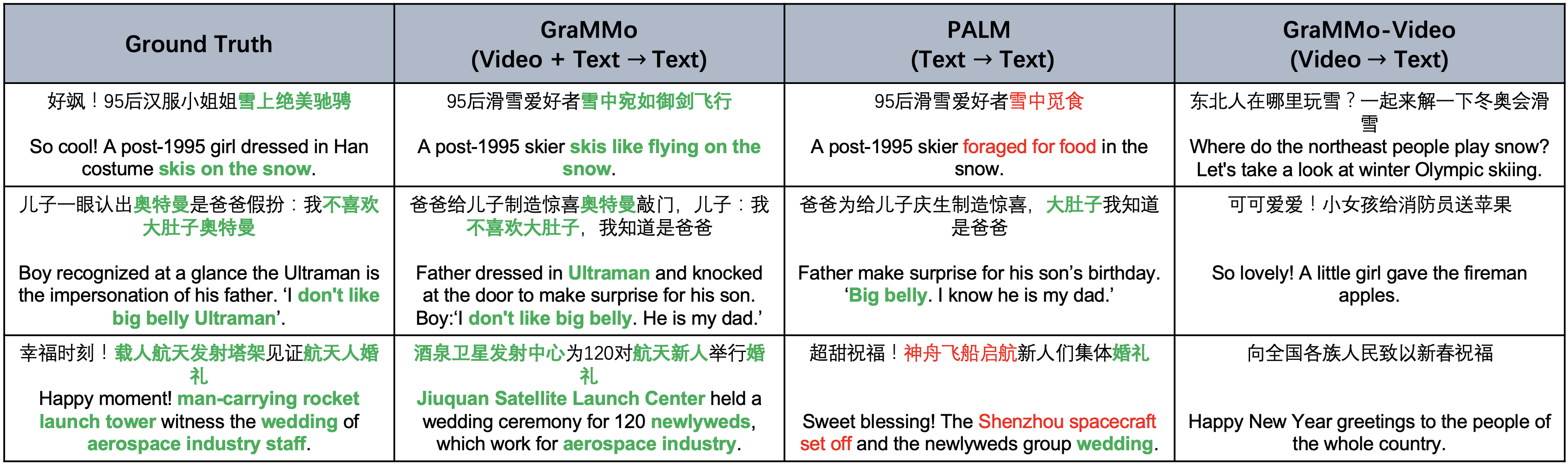}
  \caption{Case study of proposed GraMMo, PALM and GraMMo-Video.}
  \label{FIG: case study}
  \vspace{-2ex}
\end{figure*}

\subsection{Ablation Study}
To verify the contributions of each component of GraMMo, we design a series of ablation experiments. 
The results are shown in Table \ref{table:summarization result 3}. First, all summary metrics decrease when DFS and video Transformer are removed from the video encoder. The probable reasons are that DFS can improve the generalization of video embedding and video Transformer can model the sequential character of video tokens. 
Second, different fusion strategies are studied by substituting joint-modality layer. Naive concatenate and cross attention are adopted and decrease the summarization performance to great extent. 
The phenomenon illustrates that the proposed joint-modality layer is more effective in utilizing the multimodal information.


\begin{table}[t]
\small
\centering 
\begin{tabular}{p{3.5cm} | c | c | c } 
\hline 
\hline
\textbf{Methods} & \textbf{R-L} & \textbf{B-4} & \textbf{M}\\ 
\hline 
\hline
GraMMo & 35.38 & 19.65 & 35.30 \\ 
\hline
\hline
\rowcolor{green!20!blue!10}\emph{Encoder} & & & \\
w/o DFS & -0.28 & -0.13 & -0.53 \\
w/o Video Transformer & -0.98 & -0.26 & -1.03 \\
\hline
\hline
\rowcolor{green!20!blue!10}\emph{Fusion} & & & \\
Naive Concat & -0.89 & -0.96 & -1.40 \\ 
Cross Attention & -0.85 & -0.66 & -1.16 \\ 
\hline 
\hline
\end{tabular}
\caption{Ablation study on model components.} 
\label{table:summarization result 3} 
\vspace{-2ex}
\end{table}



\section{Discussion}
\subsection{Video-Text Matching Helps}
One key issue of multimodal headline generation is how to map the video and text into the joint embedding space. Therefore, the alignment of these two modalities is important, which can be reflected by video-text retrieval performance.

We selected 1,400 samples of video and its title pair from WB-News to conduct video-text retrieval experiments. 
The video embeddings $e_{v}$ and text embeddings $e_{t}$ extracted from video encoder $E_{v}(\cdot)$ and text encoder $E_{t}(\cdot)$ are computed by dot product to measure the relevant scores. 
Given a query, the most related items are recalled by sorting the scores. Recall metrics (R@1, R@5, R@10) are used to measure the results. 
As shown in Table \ref{table:video text retrieval}, the R@1 achieves about 40\% and R@5 about 60\%, which means the video and text are well aligned in one common space. Moreover, when using grafted model, the retrieval results are better than the results of train-from-scratch, reflecting the superiority of the grafted model.

\begin{table}[t]
\small
\centering 
\begin{tabular}{p{3.5cm} | c | c | c} 
\hline 
\hline
\textbf{Methods} & \textbf{R@1} & \textbf{R@5} & \textbf{R@10}\\ 
\hline 
\hline
\rowcolor{green!20!blue!10}\emph{Train from Scratch} & & & \\
text-to-video & 34.40 & 53.19 & 60.28 \\ 
video-to-text & 35.39 & 52.70 & 60.21 \\
\hline
\hline
\rowcolor{green!20!blue!10}\emph{Grafted Model} & & & \\
text-to-video & 39.65 & 60.43 & 67.73 \\ 
video-to-text & 40.35 & 60.35 & 67.30 \\
\hline 
\hline
\end{tabular}
\caption{Video-Text Retrieval Results} 
\label{table:video text retrieval} 
\vspace{-2ex}
\end{table}

\subsection{Complementary Relation}
Several examples are provided to intuitively understand the effectiveness of modality balance.
As shown in Figure \ref{FIG: case study}, the generated hypotheses of GraMMo, PALM and GraMMo-Video are presented. 
The key information is emphasised by green words, while the red words mean the wrong predictions. Compared with PALM, GraMMo can extract more key information and produce more logical expressions. This effect demonstrates the value of video modality for generating more remarkable headline. 

However, the generated hypotheses of GraMMo-Video are totally inconsistent with the ground-truths because the factual information cannot be extracted solely by video modality.
In fact, GraMMo-Video can generate the words containing similar topic semantics, which means our method can establish the connection between video and text modalities.

\subsection{Human Evaluation}


We perform human evaluation from the perspectives of readability and informativeness. For all test samples, the source video, reference headlines, and generated headline are shown to a group of people for evaluation. They need to judge the two aspects of readability and informativeness by giving an integer score in the range of 1-5, with 5 being perfect. Each sample is assessed by 5 people, and the average scores are used as the final score.

\begin{table}[t]
\small
\centering 
\begin{tabular}{p{3.0cm} | c | c  } 
\hline 
\hline
\textbf{Methods} & \textbf{Read.} & \textbf{Info.} \\ 
\hline 
\hline
Ground Truth & 4.35 & 4.05  \\ 
\hline
\hline
PALM & 3.65 & 3.08 \\
MMPT & 3.70 & 3.3\\ 
\textbf{GraMMo} & 3.71 & 3.54 \\
\hline 
\hline
\end{tabular}
\caption{Human evaluation results on readability (Read.) and informativeness (Info.) of generated headlines.} 
\label{table:Human Evaluation} 
\vspace{-2ex}
\end{table}




As shown in Table \ref{table:Human Evaluation}, we find that the GraMMo performs better readability and informativeness scores compared with PALM and MMPT, demonstrating its effectiveness in generating informative headlines. For readability, all the three headline generating methods can generate quite readable language. This is because a large training corpus can make text decoder generate coherent sentences, except for the mistakes of repetition phrases and grammatical errors. For informativeness, the major problems are the fact inconsistency and incomplete key information. They will be investigated in future work to improve the quality of generated headlines.

\subsection{Media AI Platform}
\begin{figure}[ht]
  \centering
  \includegraphics[width=1.0\linewidth]{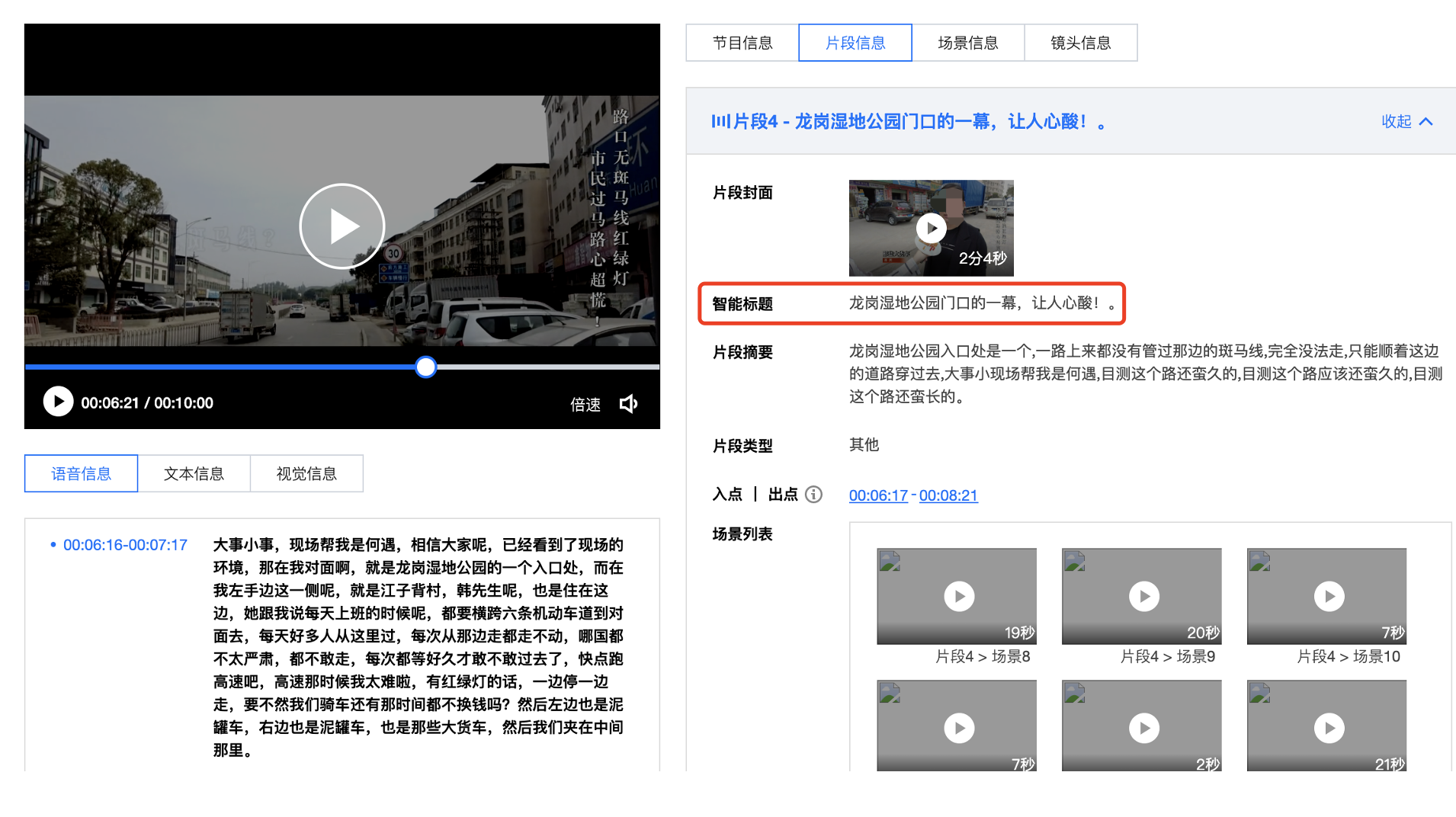}
  \includegraphics[width=1.0\linewidth]{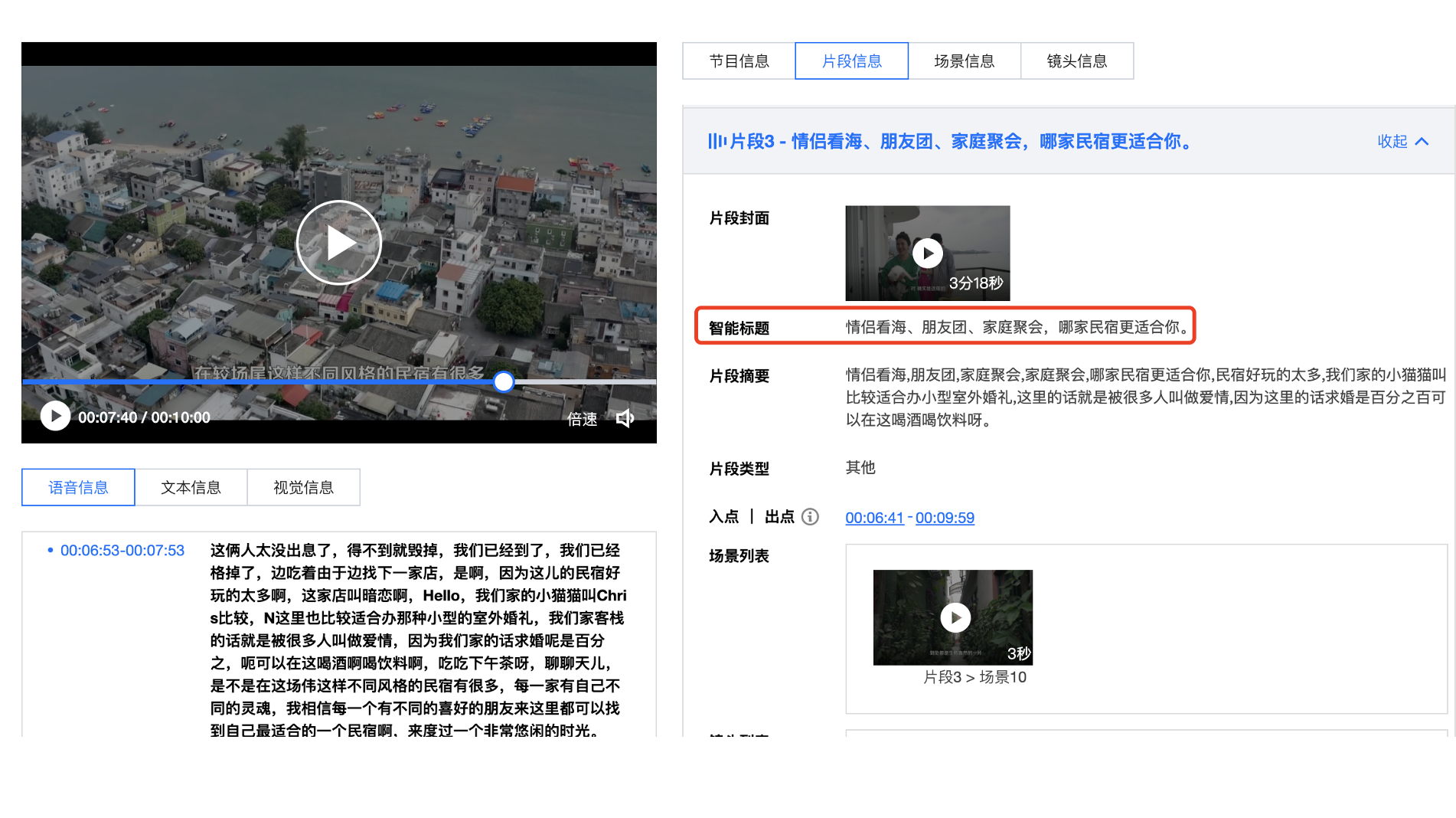}
  \includegraphics[width=1.0\linewidth]{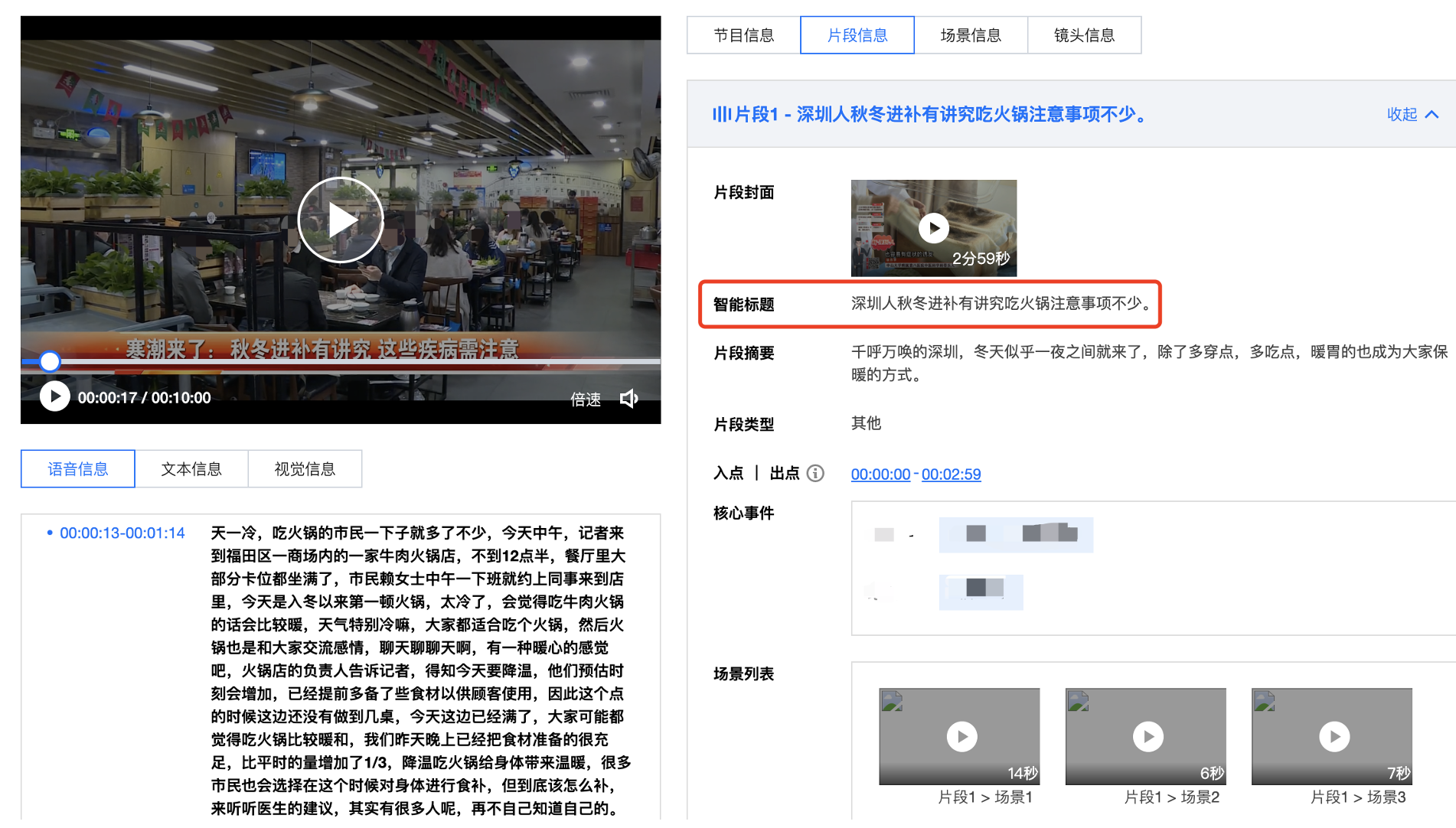}
  \caption{The examples of video headline generation results on news videos in the media AI platform. Red text boxes illustrate the generated titles based on the video and ASR text information.}
  \label{FIG: media_ai_examples}
  \vspace{-2ex}
\end{figure}

Our Chinese video headline generation is deployed in an AI platform for industrial media, which is a well-designed video understanding platform with complete video processing services. When generating headline, GraMMo takes the source video and its pre-extracted ASR text as the inputs and then predicts the textual summary as the headline of the video. We give some headline generation examples of real Chinese news videos, as shown in Figure \ref{FIG: media_ai_examples}.

\section{Conclusion}
In this paper, we propose GraMMo, grafting a pre-trained sequence-to-sequence language model and a video-language understanding model for multimodal headline generation.
By fine-tuning the representation components (video-encoder\&text-encoder) and generation component (text-decoder) of the model, we alleviate the problem of lacking large-scale dataset in multimodal headline generation.
To capture the commonality and specialty of the video and text features, we propose an extra fusion layer to balance modalities and maximally maintain the original architectures.
With this approach, we can fully take advantage of the pre-trained models, including well-trained capacity for multimodal understanding and generation.
Experiments results show that our method can significantly improve the performance and outperform similar works.
Furthermore, the proposed method has also been applied effectively and efficiently in our online system. We will release the WB-News dataset, GraMMo code, and the grafted models. 



\section{Ethics Considerations}

The authors declare that the use of data in our research is permitted. First, the Chinese corpus used in the text summarization auxiliary task is an open-source dataset. Second, for video-text data used in our multimodal headline generation, the data are collected and used in accordance with the privacy policies of short video platforms.

Ethical concerns include the usage of the proposed model for a purpose directly different from the previously mentioned headline generation task, such as hateful memes generation by feeding irrelevant video and text inputs, as well as integration in public opinion manipulation tools.

\bibliography{anthology,custom}
\bibliographystyle{acl_natbib}

\newpage

\appendix

\section{Experimental details}
\label{sec:experimental}

\subsection{Dataset}

The datasets for pre-training and fine-tuning are listed as follows.

\subsubsection{Pre-training for NLG}

We leverage 14GB high quality Chinese corpus of CLUE-small\footnote{https://www.cluebenchmarks.com/}~\cite{xu2020clue}. It contains following genres:

\paragraph{News} This sub-corpus is crawled from the We Media (self-media) platform, with a total of 3 billion Chinese words from 2.5 million news articles from roughly 63K sources.

\paragraph{WebText} With 4.1 million questions and answers, the WebText sub-corpus is crawled from Chinese Reddit-like websites such as Wukong QA, Zhihu, Sogou Wenwen, etc.

\paragraph{Wikipedia} This sub-corpus is gathered from the Chinese content on Wikipedia (Chinese Wikipedia), containing around 1.1 GB of raw texts with 0.4 billion Chinese words on a wide range of topics.

\paragraph{Comments} These comments are collected from E-commerce websites including Dianping.com and Amazon.com by SophonPlus\footnote{https://github.com/SophonPlus/ChineseNlpCorpus/}. This subset has approximately 2.3 GB of raw texts with 0.8 billion Chinese words.


\subsubsection{Pre-training for Video-Text Matching}

We collect 3.2 million videos from the Chinese video platform. The topics of video cover news, sports, entertainments, etc. For each video, its headline information is edited by uploader so that the video-text matching task can be conducted without additional manual annotations.

\subsubsection{Multimodal Generation Fine-tuning}

We establish the WB-News dataset for multimodal generation fine-tuning and evaluation. It contains more than 43,000 samples that are collected from official Weibo accounts of China's main media. 

When building WB-News, we first filter the weibo contents which have corresponding videos. Then, the raw data is manually annotated and cleaned to produce the triplet form, i.e. video, source transcript and target summary. The average video duration of these samples is about one minute, the average length of transcript is 120.6 words and the average length of summary is 21.2 words. For testing, 697 samples are selected, which can evaluate the performance of Chinese video headline generation.

\subsection{Hyper-parameters}
\label{sec:hyper-param}

\paragraph{Text Encoder}
We use PALM as the text pre-training model, in which 6-layer Transformers are used for both text encoder and decoder. The text tokens of samples are padded to 128 lengths. 

\paragraph{Video Encoder}
For the video encoder, DFS extracts 32 video tokens from each video. Then a 2-layer Transformer encoder with 8 attention heads is applied to get video embedding. 

\paragraph{Joint-modality Layer}
After obtaining the outputs encoded by text and video Transformers, linear projection layers are used to project them into the same 512 dimension. As a result, the feature dimension of $e_{v}\in R^{32\times 512}$ and $e_{t}\in R^{128\times 512}$. Then a video-text attention matrix $M_{vt}$ with 8 heads is established to produce the fusion embeddings.
\paragraph{Decoder}
In the decoding stage, we use beam search with a beam size of 5. The decoding process will not stop until an end-of-sequence (EOS) token is emitted or the length of the generated summary reaches to 64 tokens.

\paragraph{Training Details}

GraMMo is realized by fairseq toolkit\footnote{https://github.com/pytorch/fairseq}.
In training phrase, we use learning rates 3e-4 to pre-train the encoders and decoder, and 1e-5 to fine-tune the model. Batch size is set to 64 and dropout rate is 0.1. Adam optimizer is adopted with 0.01 weight-decay and 0.1 clip-norm. The training procedure runs on 8 NVIDIA
V100 GPU cards and costs about 7 days for NLG pre-training, 5 days for video-text matching pre-training and 1 hour for multimodal generation fine-tuning.

\end{document}